# Neural Network Based Reconstruction of a 3D Object from a 2D Wireframe

Kyle Johnson, Clayton Chang, and Hod Lipson

Cornell University, Ithaca, NY, USA

**ABSTRACT**

*We propose a new approach for constructing a 3D representation from a 2D wireframe drawing. A drawing is simply a parallel projection of a 3D object onto a 2D surface; humans are able to recreate mental 3D models from 2D representations very easily, yet the process is very difficult to emulate computationally. We hypothesize that our ability to perform this construction relies on the angles in the 2D scene, among other geometric properties. Being able to reproduce this reconstruction process automatically would allow for efficient and robust 3D sketch interfaces. Our research focuses on the relationship between 2D geometry observable in the sketch and 3D geometry derived from a potential 3D construction. We present a fully automated system that constructs 3D representations from 2D wireframes using a neural network in conjunction with a genetic search algorithm.*

Categories and Subject Descriptors (according to ACM CCS): I.3.6 [Computer Graphics]: Interaction techniques).

## 1. Introduction

Despite the abundance of 3D graphics software, designing 3D artifacts is still typically a cumbersome and difficult process. Raw sketching is a much more intuitive method to convey 3D information because humans are able to very easily depict and understand 3D spatial concepts on 2D medium. Because it is both quick and easy, sketching remains one of the most powerful tools used by engineers in the design stage. A sketch-based 3D reconstruction tool would allow users to maintain the simplicity of sketching while enabling them to interact with the resulting model in 3D. It should greatly enhance the user's ability to modify the design and visually communicate the results to others.

By themselves, sketches cannot be examined from different perspectives or analyzed in 3D space. A user must manually convert a sketch into a standard Computer Aided Design (CAD) model before we can analyze its shape in 3D. The ability to automatically construct a 3D object from a sketch would allow users to easily convey design concepts, as well as manipulate and modify the resulting 3D model. Humans are able to perform this 3D reconstruction with little difficulty; despite the infinite possible candidate objects, most observers readily agree on a particular representation. This consensus implies that the sketch contains some information that points humans toward a certain reconstruction. We are seeking a solution that can automatically derive what features about a particular sketch make us all visualize it in similar ways. We therefore present a novel machine learning system that can reconstruct 3D geometry from a single sketch without relying on any predefined heuristics for achieving quality reconstructions.

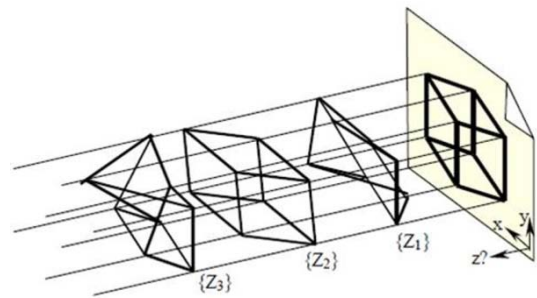

**Figure 1:** *There are an infinite number of potential reconstructions given a single sketch. We are looking for the most visually plausible.*

The crux of the system is an optimization-based algorithm that recovers vertex depths from the initial sketch. The system breaks scenes into corners formed by 3 lines in space. For each corner we feed various geometric properties as input to a neural network. The network then outputs a value that serves as the fitness function in the optimization problem. Given this fitness function, we apply a genetic algorithm to find the best solution over the set of possible reconstructions. This approach retains the flexibility of traditional optimization algorithms but does not rely on explicit geometric regularities.

We will start by describing past work on the topic, as well as our overall system design. We follow that up with a section on our reconstruction results. We conclude with a discussion of our most interesting results and the opportunities for future improvements to the system.

## 2. Related Work

A 3D wireframe undergoes an orthographic projection onto a projection plane to form the 2D sketch (see Figure 1). The projection process removes the depth information from each vertex, dropping the z coordinate. Therefore, any arbitrary set of depths assigned to the remaining (x,y) vertices constitutes a 3D object that will match the 2D projection. The task is to recover the lost depth information and determine z values that represent a valid 3D reconstruction. The human visual system is so good at interpreting sketches that we do not even realize other interpretations are possible. The reconstruction system's goal is to output depths that correspond to 3D shapes in agreement with human interpretation. Researchers have developed a variety of sketch-based 3D modeling systems in various domains [ZHH96][KS06] [BCCD04] [YSP05]. Several works have investigated the process of reconstructing 3D objects from a 2D projected sketch. Interpreting line drawings of 3D polygonal objects is a problem that has attracted considerable interest in the past [LF92][BT93][EHBE97][KHD95].

Optimization-based systems determine depth information by searching the space of possible reconstructions to minimize a fitness function. Several different fitness functions incorporating both heuristic and analytical measures have been proposed in literature, such as the minimum standard deviation of angles principle (MSDA) [Mar91], line parallelism, face planarity, corner orthogonality, and other image regularities. 2D sketches are converted to vertex/line graphs that are analyzed for these measures, and then weighted according to the probability that they correspond to 3D regularities to produce a fitness function. An overview of these reconstruction techniques can be found in [LS96] [LS00] [ML05]. There are also statistical approaches to optimization-based reconstruction [LS02]. The correlation between angles in the 2D sketch and angles in the 3D reconstruction are analyzed from a number of 3D shapes. These correlations are then used to generate probability distribution functions and reconstruct the 3D shape most likely to correspond to a set of 2D angles. Optimization-based approaches to 3D reconstruction were used by Shesh et al. [SC04] in conjunction with incremental shape construction methods

There are also several approaches not based on optimization. Huffman and Clowes [Huf71][Clo71] use line labeling to extract information about the 3D shape from the 2D sketch. Kang, et al [KML04] presented an approach where a 3D axis system for the sketch is derived from the angular distribution of the lines. This axis system is then used to determine the vertex depths by propagation along a spanning tree generated from the sketch's connectivity graph. Other methods analyze the relationship between the slopes of lines in 2D and the gradients of faces in 3D [Mac73] [Wei87].

## 3. Reconstruction

A sketch is a known set of lines and vertices in 2D. In this work we assume all edges of a sketch are straight lines and that the sketch vertices are connected, i.e. that a path can be constructed from each vertex to every other vertex. The system completes the reconstruction from a single 2D sketch without additional input or interaction from a user. Given that the 2D vertex positions (x,y) in the sketch plane are known from the sketch, the reconstruction problem consists of assigning values along the z axis to each vertex in such a way that the reconstructed shape is plausible to the eyes of human observers.

The problem of reconstructing a scene can be broken into two sub-problems: evaluating a candidate reconstruction and searching through the space of possible reconstructions. Section 3.1 describes the neural network used to evaluate candidate reconstructions. Section 3.2 describes our search strategy. In order to evaluate arbitrary scenes with a neural network with a constant number of inputs, the system breaks the sketch into corners formed by 3 lines in space. It then uses the sum of the scores of each corner outputted by the neural network over the entire scene as the value to be optimized.

### 3.1 Neural Network

#### 3.1.1 Architecture

Converting sketches into a representation consumable by the neural network was an immediate concern. The network requires a fixed number of inputs, but we wanted to handle arbitrarily complex wireframes with a single network. To address this concern, arbitrary scenes needed to be broken down into constant size parts. Then the network's output on each part needed to be recombined to compute a score for the entire object. Our approach is to split the scene into corners, formed at the intersection of 3 different edges. The network then scores individual corners of the object, rather than the object itself. The fitness score of the overall object is simply the sum of the network outputs over each corner. This implicit representation of shape puts additional onus on the network to identify crucial relationships in the data; however it is necessary to process complex objects.

The goal of the network is to predict the error of the input corner. Hence the sole network output is an estimate of the input corner's displacement in the z direction. Perfect corner projections should have an output of 0.

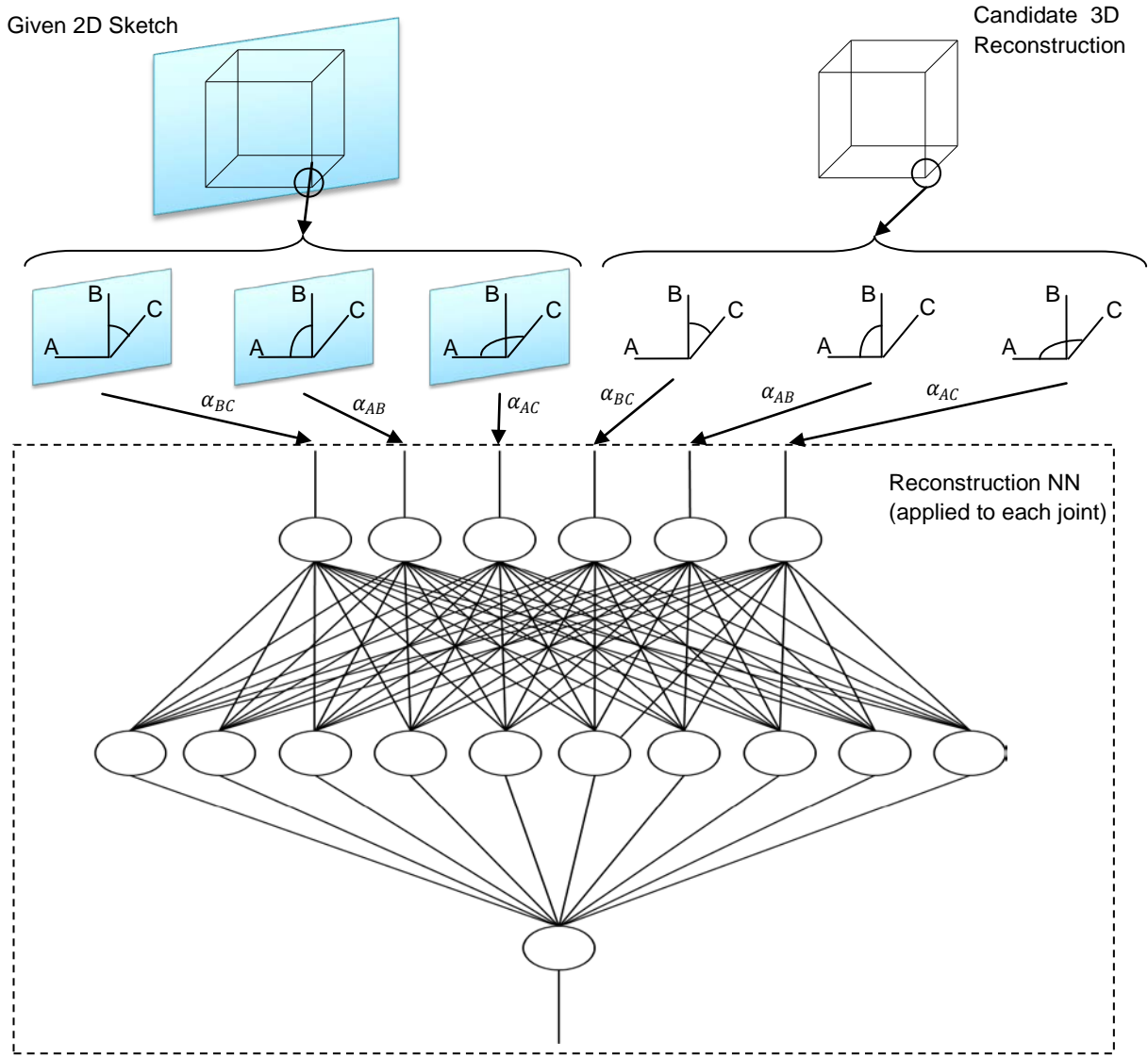

**Figure 2:** *An overview of the entire system. Six angles are computed, three from the 2D sketch and three from a candidate reconstruction. Additional inputs are used for length ratios, volumes, and areas.*

*3.1.2 Corner features*

| Feature | Description |
|---|---|
| Angles | Three 3d angles project to three 2d angles. All 6 are included. |
| Length ratios | Edges A,B,C in 3d project to a,b,c in 2d. Then A/B, A/C, B/C, a/b, a/c, and b/c make up this feature. |
| Volumes / Areas | Taking the triple product of the corner edges gives the volume of a parallelepiped based at the corner. 2d cross products give the area of the corresponding parallelogram. |

**Table 1:** *A number of features are computed for each corner. See Figure 3 as well.*

There are three general sets of features computed for each corner which are used as input to the network: angles, length ratios, and volumes. Table 1 describes each of these features. All the features have a 3D component and a 2D component. Each corner consists of a vertex v and three edges, A, B, and C. Figure 3 labels the different components of interest. Every corner has 6 angles to consider: 3 in 3D ($\alpha$, $\beta$, and $\theta$) and 3 in the 2D projection ($\alpha'$, $\beta'$, and $\theta'$). Similarly, there are 6 lengths associated with a corner: A, B, and C in 3D and a, b, and c in the projection. The ratios A/B, A/C, B/C, and, similarly, a/b, a/c, and b/c form the second feature set. Finally, the last feature consists of the volume of the tetrahedron formed by the corner vertices, plus the areas of the 3 projected triangles.

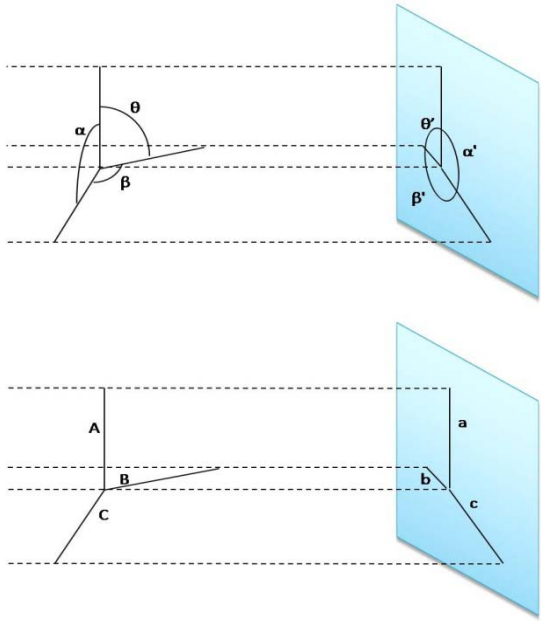

**Figure 3:** *Each corner contributes 6 angles and 6 length ratios to the network.*

### 3.1.3 Training

Randomly oriented unit triangular prisms were generated to train the neural network, such as those in Figure 4. Each corner was added to the training set with a target score of 0, since these corners have no error. Each corner is then displaced along the z-axis by x units, |x| < 0.5 (half the longest possible side length). After re-computing the features, the displaced corner is added to the training set with a target score of |x|. The displacement process is repeated for various values of x.

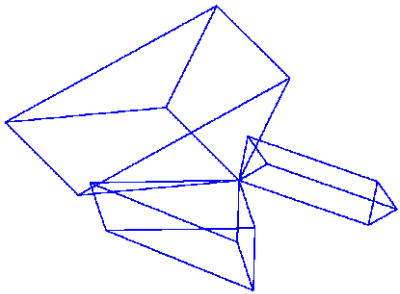

**Figure 4:** *Samples of randomly generated triangular prisms in training the neural network*

Once data was generated, the network was trained using a conjugate gradient method. This method resulted in accurate networks in very short training periods. We use early stopping to avoid overtraining our network. All the networks relied on 1000 generated prisms, training on a total of 126,000 corners.

### 3.2 Genetic Algorithm

The reconstruction is carried out by a genetic algorithm. The initial population consists of randomly generated vectors of depth assignments for each vertex. The fitness function F is simply the sum over each vertex of the neural network output.

$$F(x) = \Sigma_v \text{ net\_score}(v)$$

The reconstruction starts by calculating a fitness score for each individual in the initial population. After the population is scored, the best scoring 50% are selected and allowed to reproduce, thereby creating the next generation of depths. When reproducing, both crossovers and mutations are employed. Crossovers are implemented by drawing a random line through each individual and crossing the resulting vertex partitions. The only mutation is negation; that is, each vertex depth is negated with probably 0.5. This reconstruction process continues for a fixed number of generations. After the generations complete, a local hill-climber optimizes the best performing individual using outputs from the neural network.

### 4. Results

We tested our reconstruction algorithm on a number of generated wireframe objects of varying complexity. Our performance on the different shapes is summarized graphically in Table 2. Note that because the original 2D projection has no predefined depth axis, we are only concerned with the relative depth values of our reconstruction. Shifting all depths equally does not harm the reconstruction.

Using a network trained solely on triangular prisms of varying orientations, we were able to accurately reconstruct prisms, cubes, and some of the more complex shapes. In all cases, our final reconstruction had network error estimates very close to the estimates for the target reconstruction. For simple shapes, the predicted error approaches 0 within a few dozen generations. However more complex figures take significantly longer to evolve, usually several hundred generations or more simply to reach the error of the target shape. Even then, convergence to a visually accurate shape is not guaranteed. Because the network has no knowledge of the entire object, the reconstruction algorithm can be fooled by simple mutations of the target as simple as inverting the depths of two vertices (Figure 5).

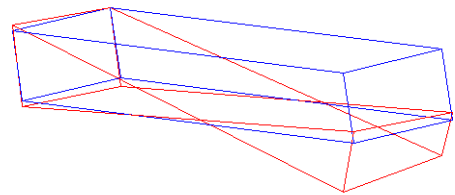

**Figure 5:** *Individual corners can all appear valid, but when put together form an invalid shape.*

In addition to the rate of convergence, the rate of successful reconstructions decreases as the complexity of the shape increases. The algorithm is able to reconstruct simple shapes nearly 100% of the time. More complex shapes, such as those at the bottom of Table 2, are more difficult to perfectly reproduce.

| 2d projection | Target 3d reconstruction | 3d reconstruction | 3d reconstruction (alternate view) | Generations/Pop |
|---|---|---|---|---|
| 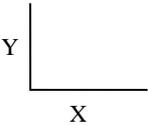 | 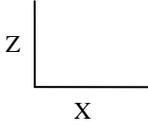 | 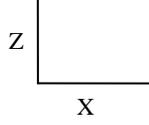 | 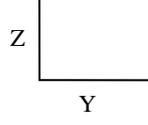 | **107 / 1000** |
| 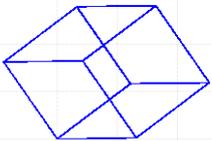 | 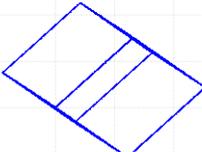 | 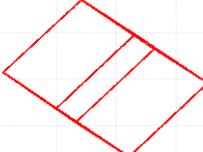 | 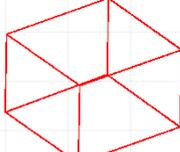 | **59 / 1000** |
| 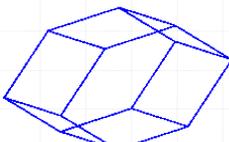 | 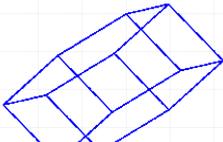 | 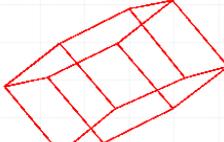 | 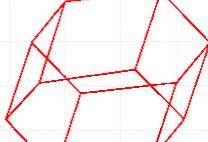 | **174 / 2000** |
| 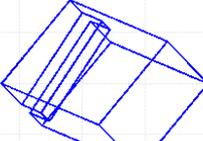 | 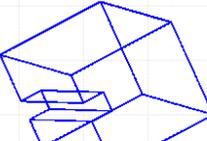 | 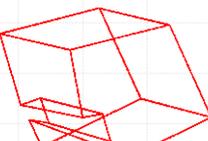 | 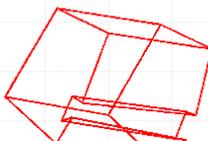 | **242 / 2500** |
| 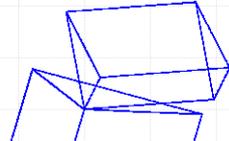 | 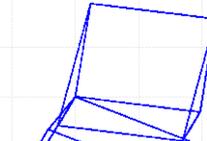 | 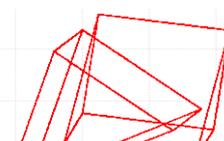 | 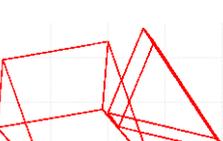 | **139 / 1000** |

**Table 2:** *A sample of reconstructions. Algorithm began with the 2d projection and generated the 3d reconstruction. Tests are auto-generated, so also included are the target 3d reconstructions. The right column shows the number of generations and population size required to perform the reconstruction.*

However, we have observed increased performance with both larger populations and more generations, and have been encouraged by the algorithm's ability to reduce the error with increased generations. Though it may not always find the global optima, it consistently avoids getting stuck in local optima. Figure 6 shows the performance of the genetic algorithm with various population sizes. The baseline is a simple hill-climbing implementation that uses the network to optimize the input projection with all depths set to 0.

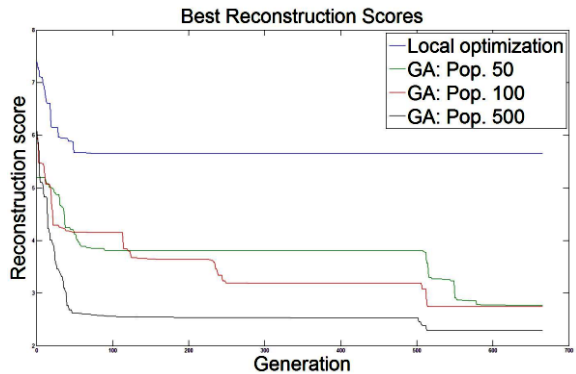

**Figure 6:** *Reconstruction performance for various algorithms, using the final shape in* **Table 2**. *Performance of the genetic algorithm increases with population size, though at the cost of speed.*

## 5. Conclusions

We have presented a machine learning based approach for reconstructing a 3D object from a single 2D wireframe with known vertices and edges. The reconstruction problem entails assigning a depth value to each vertex in the scene. To achieve this, a neural network is used to assess the validity of each vertex by estimating the error of the assigned depth. We use a genetic algorithm to search the space of possible reconstructions, with the neural network acting as a fitness function for the population. This system has successfully reconstructed a variety of scenes, ranging from simple prisms and cubes to much larger figures.

The approach is not yet as fast as other approaches previously set forth, but it also avoids relying on any predefined notions of a valid reconstruction, which was one of our primary goals. We have laid the groundwork for a reconstruction system which can learn the relationships between a 3D object and its orthographic projection. This approach may even be able to shed light on how humans mentally perform 3D reconstructions.

There is ample room for future improvement to this approach, mostly in the reconstruction process. The genetic algorithm is capable of asymptotically reducing the estimated error of the reconstruction, but this does not always result in the desired reconstruction. Given the highly fractal nature of the reconstruction space, a more intelligent and efficient genetic algorithm is needed to improve the reconstruction success rate. Reducing the number of individuals in the population that need to be scored each generation, either by ruling them out before explicitly calculating their fitness or reducing the initial population size, is integral to real-time solution. Future work will focus on increasing the efficiency of reconstruction.

Outside of efficiency, there is work to be done improving the robustness of the algorithm. Currently the system has only been tested on single-object scenes. More research is needed on dealing with multi-object scenes; reconstructing individual objects separately in parallel requires an intelligent method of combining the resulting reconstructions so as the whole scene appears plausible. In a single-object scene, the algorithm is only concerned with the relative depths of the vertices; the scale of the z-axis is inconsequential. Once multiple objects are introduced into the scene, the actual depths of each reconstruction become relevant, adding significant complexity to the overall reconstruction process.

## 6. Acknowledgements

This work was supported in part by the U.S. National Science Foundation under grant ISS-0428133.